%% file: main.tex
\ifdefined\XeTeXversion\else\ifdefined\pdfoutput\pdfoutput=1\fi\fi
\providecommand{\arxivroot}{}
\documentclass[10pt, logo, twocolumn, copyright]{nvidiatechreport}
\newcommand{\stok}[1]{\colorbox{blue!12}{\strut\texttt{#1}}}
\newcommand{\ttok}[1]{\colorbox{orange!20}{\strut\texttt{#1}}}
\usepackage{caption}
\usepackage{mdframed}
\usepackage[utf8]{inputenc} 
\usepackage[T1]{fontenc}    
\usepackage{hyperref}       
\usepackage{url}            

\usepackage{booktabs}       
\usepackage{amsfonts}       
\usepackage{nicefrac}       
\usepackage{microtype}      
\usepackage[dvipsnames]{xcolor}         
\usepackage{multirow}
\usepackage{multicol}
\usepackage{graphicx}
\usepackage{subcaption}
\usepackage[numbers]{natbib}
\usepackage{tabto}
\usepackage{xspace}
\usepackage{amsmath}
\usepackage{adjustbox}
\usepackage{enumitem}
\usepackage{wrapfig}
\usepackage{dblfloatfix}
\usepackage{algorithm}
\usepackage{algpseudocode}
\usepackage{comment}
\usepackage{cuted}
\usepackage[most]{tcolorbox}
\usepackage{xcolor}
\usepackage{graphicx}
\usepackage{hyperref}
\usepackage{makecell}
\usepackage{pifont}
\usepackage{amsthm}

\newtheorem{proposition}{Proposition}

\newif\ifshowchanges
\showchangesfalse
\definecolor{revisiongreen}{RGB}{0,120,45}
\DeclareRobustCommand{\rev}[1]{{\ifshowchanges\color{revisiongreen}\hypersetup{allcolors=revisiongreen}\fi #1}}
\newenvironment{revision}{\begingroup\ifshowchanges\color{revisiongreen}\hypersetup{allcolors=revisiongreen}\captionsetup{labelfont={color=revisiongreen},textfont={color=revisiongreen}}\renewcommand{\headingfont}{\color{revisiongreen}\bfseries\fontsize{13}{14}\selectfont}\fi}{\par\endgroup}
\titleformat{\section}{\large\bfseries\headingfont\raggedright}{\rev{\thesection.}}{0.5em}{#1}[]
\titleformat{\subsection}{\bfseries}{\rev{\thesubsection.}}{0.5em}{#1}[]
\titleformat{\subsubsection}{\bfseries\itshape}{\rev{\thesubsubsection.}}{0.5em}{#1}[]
\graphicspath{{\arxivroot}{../}}
\usepackage{etoolbox}
\csdef{reviewbib@chen2025cdm}{}
\csdef{reviewbib@cui2025multilevel}{}
\csdef{reviewbib@wang2026sedi}{}
\csdef{reviewbib@li2025tokalign}{}
\csdef{reviewbib@gao2021framework}{}
\csdef{reviewbib@lee2025genrecal}{}
\newcommand{\revisionbibcolor}[1]{\ifshowchanges\ifcsdef{reviewbib@#1}{\color{revisiongreen}}{}\fi}
\AtBeginDocument{%
 \let\originalbibitem\bibitem
 \renewcommand{\bibitem}[2][]{\par\normalcolor\revisionbibcolor{#2}\ifstrempty{#1}{\originalbibitem{#2}}{\originalbibitem[#1]{#2}}\revisionbibcolor{#2}}%
}

\newcommand{\cmark}{\ding{51}}
\newcommand{\xmark}{\ding{55}}
\newcolumntype{g}{>{\color{gray}}c} 

\title{X-Token: Projection-Guided Cross-Tokenizer Knowledge Distillation}
\author{
Sharath Turuvekere Sreenivas*,
Adithyakrishna Venkatesh Hanasoge*,
Mingyu Yang*,
Ali Taghibakhshi,
Saurav Muralidharan,
Ashwath Aithal,
Pavlo Molchanov
}

\correspondingauthor{sharatht@nvidia.com}

\begin{abstract}
\textbf{Abstract:}
\rev{Cross-tokenizer knowledge distillation allows a student to learn from teachers with incompatible vocabularies while preserving its tokenizer and inference architecture. Existing logit-based methods either supervise only the observed token or match full distributions through strict token partitions and heuristic ranking. We identify two limitations: an \emph{uncommon-token failure}, where task-critical tokens outside the shared vocabulary are suppressed, and \emph{over-conservative matching}, where exact string equality discards useful correspondences. We propose \textbf{X-Token}, which addresses both through a sparse projection initialized from tokenizer-level string rules and optionally refined during distillation. \textbf{P-KL} removes the partition and projects the student distribution into the teacher vocabulary; \textbf{H-KL} retains the hybrid loss while relaxing matching through the projection. A token-coverage audit guides the choice between them, and both extend naturally to heterogeneous teachers. Across Llama, OLMo, and Gemma students, X-Token achieves the highest average accuracy in all four teacher--student settings among all evaluated cross-tokenizer methods. Complementary teachers provide a further 1.30-point gain, and both variants are faster than the evaluated cross-tokenizer baselines in matched profiling.}
\end{abstract}

\ifdefined\XeTeXversion\microtypesetup{tracking=false}\fi
\begin{document}
\maketitle


\input{tex/introduction.tex}
\input{tex/preliminaries.tex}
\input{tex/method.tex}

\input{tex/experiment.tex}
\input{tex/related.tex}

\section*{Conclusions}

\begin{revision}
X-Token addresses harmful token partitioning and overly strict matching through a sparse tokenizer-derived projection and two complementary objectives. Coverage-guided selection of P-KL or H-KL yields the highest average accuracy against all evaluated cross-tokenizer baselines across three student families, while complementary teachers provide further gains. Both variants outperform the evaluated cross-tokenizer baselines in matched throughput measurements.

\paragraph{Limitations and future work.}
Loss selection requires manually specified task-critical categories; our validation focuses on numerals and small base-model students. Broader multilingual and post-training evaluations, automatic category discovery, and teacher selection remain future work. Seed analysis is limited to the Phi comparison, and runtime depends on the implementation and hardware.

\paragraph{Implementation availability.}
X-Token is publicly available in \href{https://github.com/NVIDIA-NeMo/RL/blob/main/docs/guides/xtoken-off-policy-distillation.md}{NeMo RL}, including projection-construction tools and single- and multi-teacher distillation recipes.
\end{revision}

\section{Acknowledgments}
We would like to thank our colleagues and leaders at NVIDIA for their valuable support and feedback. We are especially grateful to Shizhe Diao for assistance with datasets, and to Marcin Chochowski, Sepehr Sameni, and Daniel Korzekwa for their insightful discussions and constructive feedback. \rev{We also thank Byung-Kwan Lee for valuable discussions drawing on ideas in GenRecal~\citep{lee2025genrecal} and their connections to cross-tokenizer and cross-model knowledge distillation.}

\clearpage
{
  \small
  \bibliographystyle{unsrt}
  \bibliography{paper}
}

\clearpage
\appendix
\input{tex/appendix.tex}
\input{tex/review_appendix.tex}

\end{document}

%% file: tex/introduction.tex
\section{Introduction}
\label{sec:intro}

Knowledge distillation (KD)~\citep{hinton2015distilling, romero2014fitnets, furlanello2018born, park2019relational} transfers the `dark knowledge' in a teacher's output distribution to a student, typically via per-position Kullback--Leibler (KL) divergence over next-token probability distribution. This formulation requires a shared tokenizer, effectively tying the student to same-family teachers. As a result, a practitioner committed to a given tokenizer (\emph{e.g.}, Llama-3.2-1B \cite{grattafiori2024llama}) cannot leverage stronger or more specialized teachers with incompatible tokenizers (\emph{e.g.}, Phi-4-mini \cite{abouelenin2025phi}, Qwen3-4B\rev{-Base} \cite{yang2025qwen3}). This constraint also prevents multi-teacher distillation across tokenizer families, limiting the ability to combine teachers with complementary strengths (\emph{e.g.}, reasoning, code, multilingual) into a unified training signal. Cross-tokenizer distillation removes this restriction, enabling both freedom from teacher-tokenizer lock-in and effective multi-teacher learning from diverse sources.

\begin{figure*}[t]
  \centering
  \includegraphics[width=\linewidth]{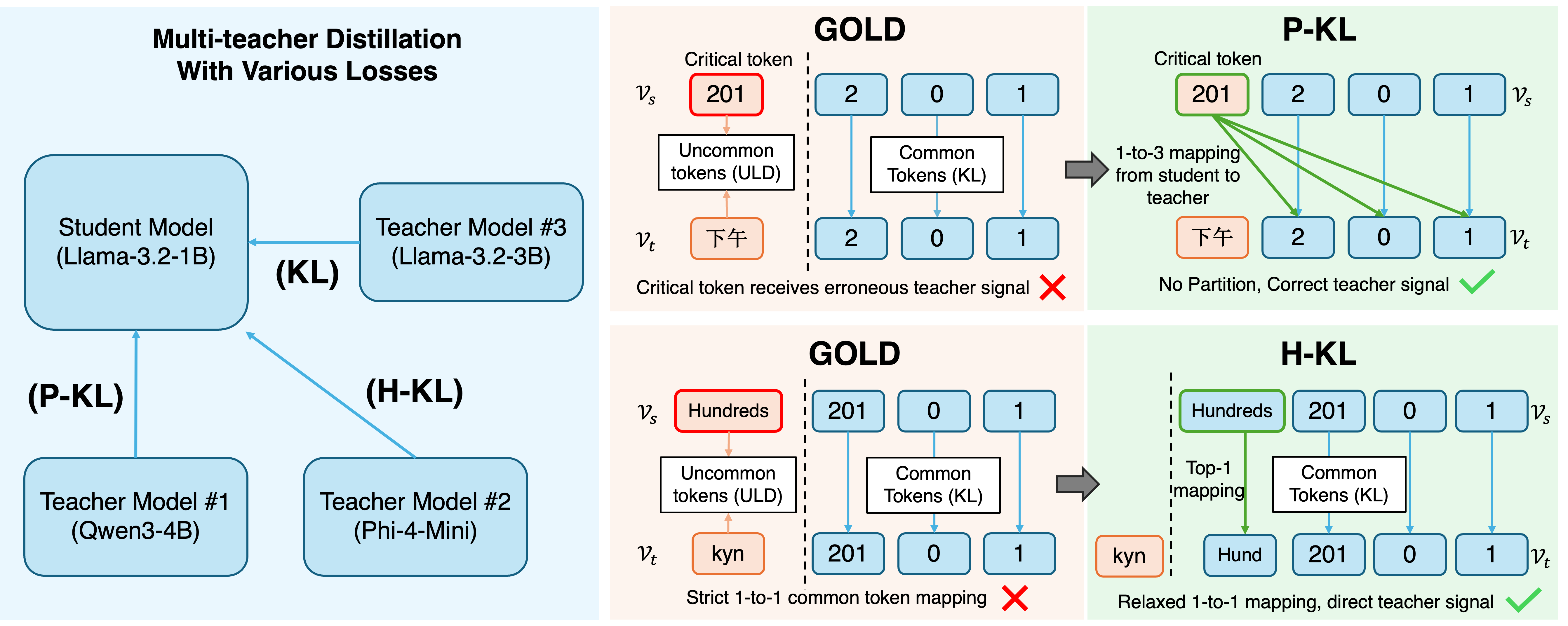}
  \vspace{-0.2in}
  \caption{\textbf{Left:} multi-teacher distillation routes each
    teacher through its appropriate loss --- KL for the
    same-tokenizer Llama-3.2-3B, P-KL/H-KL for cross-tokenizer
    Qwen3-4B\rev{-Base} and Phi-4-mini.
  \textbf{Right:} X-Token addresses two failure modes of GOLD's
    string-equality partition and composes across teachers.
    \textbf{Right, top:} the critical token \texttt{201} has no
    GOLD match and receives erroneous signal; \textbf{P-KL}
    connects it to $\{2,0,1\}$ in $\mathcal{V}_T$ via the
    projection $W$.  \textbf{Right, bottom:} \texttt{Hundreds} is
    excluded from GOLD's common-KL term; \textbf{H-KL} admits
    $(\texttt{Hundreds},\texttt{Hund})$ via the top-$1$ of $W$.
    }
  \label{fig:overview}
  \vspace{-0.1in}
\end{figure*}

Existing cross-tokenizer KD methods fall into two broad families: \emph{representation-based} approaches that align the teacher and student at the embedding or hidden-state level (\emph{e.g.}, DSKD~\cite{zhang2024dual}), and \emph{logit-distribution-based} approaches that operate directly on output distributions (\emph{e.g.}, ULD~\citep{boizard2024towards}, GOLD~\citep{patiño2025_unlocking_on_policy_distillation_for_any_model_family}). The latter approaches are particularly attractive at continual-pretraining scale, \rev{as they can preserve the student tokenizer and inference architecture while changing the distillation objective.} 

GOLD~\citep{patiño2025_unlocking_on_policy_distillation_for_any_model_family} applies a hybrid loss that partitions tokens into a 1-to-1 string-matched \emph{common} subset trained with KL divergence and an \emph{uncommon} remainder matched via rank-based L1 following ULD \citep{boizard2024towards}. However, this hybrid design exhibits two structural limitations. First, an \emph{uncommon-token failure}: when tokenizers fragment differently (\emph{e.g.}, Qwen3 splits multi-digit numerals while Llama-3 packs them as single tokens), \emph{critical tokens}---tokens whose correct prediction directly determines task accuracy (\emph{e.g.}, multi-digit numerals in GSM8k)---are forced into the unmatched subset. These tokens are then degraded by (i) identity-agnostic noise from rank-based matching and (ii) suppressive gradients from the common-KL term acting through the full-vocabulary softmax. Second, \emph{over-conservative matching}: strict string-equality excludes equivalent token pairs across tokenizers, both in surface form and over teacher multi-token decompositions---leaving useful alignment signal unexploited even when the partition is otherwise well-formed (\emph{e.g.}, a student token \texttt{Hundreds} corresponds to teacher tokens \texttt{Hund} followed by \texttt{reds}, but strict matching discards this correspondence due to lack of exact string equality).

We propose \textbf{X-Token} (Figure~\ref{fig:overview}), which addresses the limitations of current methods and makes the following contributions:
\begin{revision}
\begin{itemize}
\item \textbf{Deterministic cross-tokenizer alignment.}
We combine scalable DP-based sequence alignment with a sparse projection $W$ built from canonicalized string matching and multi-token decoding rules; $W$ can be refined jointly during KD.

\item \textbf{Complementary objectives and coverage-guided selection.}
\textbf{P-KL} removes harmful partitioning, while \textbf{H-KL} relaxes matching through top-ranked mappings under $W$. A coverage audit of task-critical token categories guides objective selection.

\item \textbf{Multi-teacher KD across tokenizer families.}
X-Token combines heterogeneous teachers, with complementary strengths yielding \textbf{+1.3} points over the best single teacher. Simple static weighting outperforms the tested adaptive schemes.

\item \rev{\textbf{Accuracy and efficiency across student families.}
X-Token achieves the highest average accuracy in all four teacher--student settings across Llama, OLMo, and Gemma, outperforming all evaluated cross-tokenizer baselines. Both variants are faster than the evaluated cross-tokenizer baselines in matched profiling.}
\end{itemize}
\end{revision}

%% file: tex/preliminaries.tex
\begin{revision}
\section{Preliminaries}
\label{sec:preliminaries}
\subsection{Knowledge Distillation}
\label{sec:kd_background}

We adopt the standard KD objective~\citep{hinton2015distilling},
over distributions defined on a shared support. For $N_c$ aligned chunks, let $\hat p_S^{(k)}$ and $\hat p_T^{(k)}$ denote the normalized student and teacher distributions. When the supports agree, the objective is:
\[
\mathcal{L}_{\mathrm{KD}}
= \frac{1}{N_c}\sum_{k=1}^{N_c}
\mathrm{KL}\!\bigl(\hat{p}_T^{(k)} \;\|\; \hat{p}_S^{(k)}\bigr).
\]

\subsection{Hybrid Loss Formulation}
\label{sec:hybrid_loss_formulation}

We first formalize the partition-based hybrid loss used in
GOLD~\citep{patiño2025_unlocking_on_policy_distillation_for_any_model_family},
which serves as the baseline for our loss variants. This formulation
partitions the vocabularies into a 1-to-1 string-matched \emph{common}
subset $\mathcal{C}$ and uncommon remainders
$\mathcal{U}_S, \mathcal{U}_T$. It applies direct KL on $\mathcal{C}$
and rank-sorted $L_1$ matching on $\mathcal{U}=\mathcal{U}_S \cup \mathcal{U}_T$:
\begin{align}
  \mathcal{L}_{\mathrm{common}}^{(k)}
  &= \sum_{(s, t) \in \mathcal{C}}
       \hat{p}_T^{(k)}[t]\,
       \bigl(\log\hat{p}_T^{(k)}[t] - \log\hat{p}_S^{(k)}[s]\bigr),
       \label{eq:gold_common}\\
  \mathcal{L}_{\mathrm{ULD}}^{(k)}
  &= \Bigl\|
       \mathrm{sort}_{\downarrow}\!\bigl(\hat{p}_S^{(k)}\big|_{\mathcal{U}_S}\bigr)
       -
       \mathrm{sort}_{\downarrow}\!\bigl(\hat{p}_T^{(k)}\big|_{\mathcal{U}_T}\bigr)
     \Bigr\|_1,
       \label{eq:gold_uld}\\
  \mathcal{L}_{\mathrm{GOLD}}^{(k)}
  &= \lambda_{\mathrm{KL}}\,\mathcal{L}_{\mathrm{common}}^{(k)}
   + \lambda_{\mathrm{ULD}}\,\mathcal{L}_{\mathrm{ULD}}^{(k)}.
       \label{eq:gold_total}
\end{align}

While this hybrid formulation enables cross-tokenizer KD, it
introduces undesirable gradient behaviors on tokens in the
uncommon set, which motivates the projection-based objectives below.
\end{revision}

%% file: tex/method.tex
\section{Method}
\label{sec:method}

X-Token consists of three components: (i) span alignment to produce
text-consistent units, (ii) a projection matrix $W$ to bridge
vocabularies, and (iii) two complementary loss formulations,
\textbf{P-KL} and \textbf{H-KL}, with an optional multi-teacher
extension. Together, these enable distillation across mismatched
tokenizers.

\subsection{Sequence Alignment}
\label{sec:xtokenizer_challenge}

When teacher and student use different tokenizers
$\mathcal{T}_S, \mathcal{T}_T$, token sequences differ in length and
lack positional correspondence, making per-position KD ill-defined.

We address this via \emph{span alignment}, grouping tokens into chunks
$\{(A_k^S, A_k^T)\}_{k=1}^{K}$ that decode to the same underlying text.
We then apply a chain-rule merge over each chunk to obtain chunk-level
distributions $\hat{p}_S^{(k)}$ and $\hat{p}_T^{(k)}$, which serve as
aligned units for distillation. Such approach was inspired by~\cite{minixhofer2025universal}.

\rev{The chain-rule merge scores each retained candidate separately and normalizes the resulting scores, preserving a distribution rather than only the observed chunk's likelihood. Loss computation uses a shared top-8K teacher support (8,192 entries), distinct from the top-$4$ mappings per row of $W$.}

\subsection{X-Token Projection Matrix $W$}
\label{sec:projection}

Even after alignment, teacher and student distributions are defined
over different vocabularies. We introduce a projection matrix
$W\in\mathbb{R}^{|\mathcal{V}_S|\times|\mathcal{V}_T|}$ that maps
student-token probabilities into teacher vocabulary space, where $\mathcal{V}_S$ and $\mathcal{V}_T$ represents the student and teacher vocabularies.

\begin{figure}[t]
    \vspace{-0.2in}
    \centering
    \includegraphics[width=\linewidth]{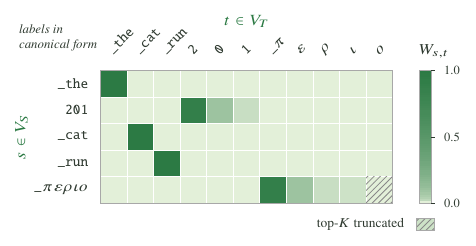}
    \vspace{-0.22in}
    \caption{Subset of the projection matrix $W$ for a Llama-3.2 student and Qwen-3 teacher. Exact matches include \texttt{\_the}, \texttt{\_cat}, and \texttt{\_run}. For tokens without exact matches, the multi-token rule is applied: \emph{e.g.}, \texttt{201}$\to(2,0,1)$, and the Greek prefix \texttt{\_$\pi\varepsilon\rho\iota o$} maps to five teacher sub-tokens, with the lower weight entries (hatched) truncated (top-$K$=4).
    }
    \label{fig:proj-matrix}
    \vspace{-0.1in}
\end{figure}
As visualized in Figure~\ref{fig:proj-matrix}, the projection matrix
$W\in\mathbb{R}^{|\mathcal{V}_S|\times|\mathcal{V}_T|}$ maps each
student token to a weighted combination of teacher tokens.  We
construct $W$ deterministically in two passes.
\emph{(1)~Exact-match pass}: for every $(s,t)\in\mathcal{V}_S\times\mathcal{V}_T$ whose decoded
strings match after canonicalization (unifies space prefixes
such as \texttt{\.{G}}/\texttt{\_} and newline markers), set
$W[s,t]=1$. We use $s$ and $t$ interchangeably with their integer indices under the canonical vocabulary ordering when indexing arrays or matrices.
\emph{(2)~Multi-token-rule pass}: for each remaining student token
$s$, re-tokenize its decoded text under the teacher tokenizer to
yield a sequence $(\tau_0,\ldots,\tau_{\ell})\in\mathcal{V}_T$ (here
$\tau_i\in\mathcal{V}_T$ is the $i$-th index in this
re-tokenization), and set $W[s,\tau_i]=\beta\,\gamma^{i}$ with
$(\beta,\gamma)=(0.9,0.1)$. \rev{The full teacher decomposition is considered before retaining the top-$4$ mappings. We normalize the retained weights; after projection and restriction to the retained teacher support, we normalize again before computing KL.}  The matrix $W$ is constructed once before
training and can be optionally fine-tuned during KD; full
pseudocode is provided in Appendix~\ref{app:projection}.

\begin{revision}
With at most four entries per student token, $W$ requires $O(|\mathcal V_S|)$ storage; learning updates only its retained values. Strong positional decay emphasizes the first teacher sub-token, which is predicted directly from the aligned prefix. This is an empirical initialization, evaluated in Appendix~\ref{app:projection_ablations}.
\end{revision}

\subsection{P-KL: Addressing Erroneous and Suppressive Gradients in Hybrid Loss}
\label{sec:pkl}

GOLD's hybrid loss induces two undesirable gradient behaviors on
uncommon student logits (Figure~\ref{fig:overview}).

\textbf{Erroneous gradients from rank-based matching:}
the ULD term $\mathcal{L}_{\mathrm{ULD}}$ matches tokens in the
uncommon set by rank, pairing each student token with a teacher
token of similar rank rather than semantic correspondence. This produces identity-agnostic gradients that misalign critical tokens (\emph{e.g.}, numerals) with unrelated teacher tokens (\emph{e.g.}, special characters), degrading supervision quality.

\textbf{Suppressive gradients from the common-KL term:}
the common-KL term $\mathcal{L}_{\mathrm{common}}$
(Eq.~\ref{eq:gold_common}) is computed using full-vocabulary
softmax. Although uncommon tokens do not appear explicitly in the loss,
the normalization induces gradients on all logits, reducing the
relative probability of tokens in $\mathcal{U}$. Detailed proof can be found in the Appendix.

Together, these effects yield weak or misdirected supervision for
uncommon tokens, particularly when critical tokens fall into
$\mathcal{U}$ (\emph{e.g.}, Llama's $1{,}100$ multi-digit numerals under
a digit-splitting Qwen tokenizer), leading to degraded performance. (\emph{e.g.},
GSM8k drops to $2.56$ vs.\ $12.89$ for same-tokenizer KD from a
weaker teacher).

To address this limitation, we leverage \textbf{P-KL}, which projects the student distribution $\hat{p}_S^{(k)}$ into teacher vocabulary space $\tilde{p}_S^{(k)}$, enabling direct alignment with the teacher distribution (Figure~\ref{fig:overview}). Here, $s$ and $t$ index student and teacher tokens:
\begin{revision}
\begin{equation}
 \begin{aligned}
 \tilde p_S^{(k)}[t]
 &=\frac{\sum_s W[s,t]\hat p_S^{(k)}[s]}
 {\sum_{u\in\mathcal R_T}\sum_s W[s,u]\hat p_S^{(k)}[s]},\\
 \mathcal L_P^{(k)}
 &=\mathrm{KL}\!\bigl(\hat p_T^{(k)}\,\|\,\tilde p_S^{(k)}\bigr).
 \end{aligned}
 \label{eq:pkl}
\end{equation}
\end{revision}
\rev{Here $t\in\mathcal R_T$, and both distributions in the KL are normalized over the retained teacher support $\mathcal R_T$.}

This formulation replaces both sources of error with
teacher-aware supervision over all tokens, including those in
$\mathcal{U}$ (\emph{e.g.},
\texttt{201} onto $[2,0,1]$), by directly aligning the student distribution with
the teacher distribution, restoring the guidance the partition discards.

\subsection{H-KL: Relaxing the 1-to-1 Matching}
\label{sec:hkl}

When no critical token is routed into $\mathcal{U}$ --- a
condition we audit per category on the student vocabulary
(Table~\ref{tab:audit_qwen_phi}) --- the partition itself is a
useful feature: the common-KL on identity-aligned pairs delivers
per-pair KL targets that are sharper than P-KL's projection,
which blends student probability mass across multiple teacher
tokens through the multi-token-rule rows of $W$.  The opportunity
here is not to drop the partition but to make it less wasteful:
GOLD's string-equality criterion is conservative compared to the
richer set of sub-token matches that $W$ exposes through
teacher-side re-tokenization of the student's decoded text.
Table~\ref{tab:loss_variant} confirms this empirically:
H-KL outperforms P-KL by $+1.68$ avg.\ on the Phi-4-mini teacher
where this precondition holds.

To address this, \textbf{H-KL} retains the hybrid structure but
relaxes the definition of $\mathcal{C}$ using the projection
matrix $W$. For each student token $s$, we select its
top-ranked teacher token $t^*$:
\begin{equation}
  t^* = \arg\max_{t'\in\mathcal{V}_T}\,W[s, t'],\;\;
        W[s, t^*] > 0,
\end{equation}
and extend the common $\mathcal{C}$ set with ${(s, t^*)}$.
This construction expands $\mathcal{C}$ beyond strict string
matches by incorporating high-confidence alignments induced by
$W$. Exact matches are preserved since they receive the highest
weight in $W$, while additional near-equivalent pairs are included
when no exact correspondence exists.

H-KL then applies the hybrid loss formulation
(Eq.~\ref{eq:gold_total}) over this expanded set $\mathcal{C}$.
This improves coverage of aligned token pairs while retaining the
benefits of direct KL supervision, yielding a $+0.5$ average accuracy gain (Table~\ref{tab:main}). A pair like
$(\texttt{Hundreds},\texttt{Hund})$ that strict equality
excludes is now admitted into $\mathcal{C}$ and contributes the
same direct-KL signal as a native exact match.

\rev{On Phi-to-Llama, H-KL adds 622 pairs to GOLD's common set; Appendix~\ref{app:projection_ablations} reports seed variability.}

\subsection{Multi-Teacher Distillation}
\label{sec:multiteacher}

Given $M$ teachers indexed by $m \in \{1,\ldots,M\}$, each with its own projection matrix $W_m$ and choice of P-KL or H-KL, X-Token naturally extends to the multi-teacher distillation by aggregating per-teacher losses:
\begin{equation}
\mathcal{L}_{\mathrm{KD,multi}}
= \sum_{m=1}^{M} \alpha_m\,
  \frac{1}{|\mathcal{K}_m|}\sum_{k\in\mathcal{K}_m}
  \mathcal{L}_{\text{*},m}^{(k)}
\label{eq:multiteacher}
\end{equation}
where $\mathcal{L}_{\text{*},m}^{(k)}\in\{\mathcal{L}_P^{(k)}, \mathcal{L}_H^{(k)},\mathcal{L}_{\mathrm{KL}}^{(k)}\}$  denotes the selected loss for teacher $m$ --- P-KL or H-KL for cross-tokenizer teachers, and standard token-level KL for same-tokenizer teachers. We consider several choices for $\alpha_m$, based on cross-entropy, entropy, and maximum predicted probability, with the goal of assigning higher weight to more confident teachers. In practice, however, we find that simple static weighting performs best (Table~\ref{tab:mt_weighting}).

Beyond weighting, our results highlight that \emph{teacher complementarity} plays a critical role: combinations of teachers with diverse strengths (\emph{e.g.}, math vs.\ general knowledge) consistently outperform more homogeneous pairings (Table~\ref{tab:main}). This suggests that effective multi-teacher distillation benefits not only from how teachers are weighted, but also from which teachers are selected.

\subsection{Dynamic KD/CE Scaling}
\label{sec:dynamic_scaling}

We combine the distillation loss $\mathcal{L}_{\mathrm{KD}}$
(single teacher) or $\mathcal{L}_{\mathrm{KD,multi}}$ (multi-teacher)
with the next-token cross-entropy $\mathcal{L}_{\mathrm{CE}}$.
As these terms can differ significantly in magnitude and vary
\rev{throughout training, we rescale KD to match the scalar CE loss and avoid tuning another KD/CE coefficient for every setting:}
\begin{equation}
  \mathcal{L}
  \;=\;\mathrm{sg}\!\left(
    \frac{\mathcal{L}_{\mathrm{CE}}}{\mathcal{L}_{\mathrm{KD}}}
  \right)\cdot\mathcal{L}_{\mathrm{KD}}
   \,+\,\mathcal{L}_{\mathrm{CE}},
  \label{eq:total_loss}
\end{equation}
\begin{revision}
Here $\mathrm{sg}(\cdot)$ denotes stop-gradient. This heuristic matches scalar loss magnitudes, not gradient norms, and avoids per-setting coefficient tuning. For multiple teachers, it rescales the aggregated KD loss.
\end{revision}

\subsection{Selecting P-KL vs H-KL via Coverage Analysis}
\label{sec:choose_loss}

We select between P-KL and H-KL using a coverage-based criterion across the student and teacher vocabularies. Tokens are grouped into character classes (\emph{e.g.}, digits by length, alphabetic, punctuation, multi-byte / non-ASCII), and we measure their retention in the common set $\mathcal{C}$.

For math tasks, multi-digit numerals are critical: under Qwen3-4B\rev{-Base}, all $1{,}100$ Llama two and three-digit numerals fall into $\mathcal{U}$, whereas under Phi-4-mini-Instruct they remain in $\mathcal{C}$. In contrast, ASCII punctuation and single-digit numerals are fully covered in both cases (Table~\ref{tab:audit_qwen_phi}). Accordingly, we use \textbf{P-KL} when critical tokens fall outside $\mathcal{C}$ (Qwen) and \textbf{H-KL} otherwise (Phi-4-mini).

\begin{revision}
The audit is automatic once task-critical categories are specified, although choosing those categories requires domain knowledge. The OLMo and Gemma results validate the same numeral-based diagnostic across student families (Table~\ref{tab:matched_baselines}).
\end{revision}

\paragraph{X-Token training step (single teacher).}
Algorithm~\ref{alg:xtoken} summarizes the per-step computation.  The mode parameter
$\mathcal{M}\in\{\textsf{P-KL},\textsf{H-KL}\}$ is fixed per
teacher.

\begin{algorithm}[t]
  \caption{X-Token training step (single teacher).}
  \label{alg:xtoken}
  \begin{algorithmic}[1]
    \Require Student $f_S$ (trainable); teacher $f_T$ (frozen);
             top-$4$ projection matrix $W$ (rule-based init,
             jointly learned for P-KL); loss mode
             $\mathcal{M}\!\in\!\{\textsf{P-KL},\textsf{H-KL}\}$;
             input text $x$; temperature $\tau$; loss weights
             $\lambda_{\mathrm{KL}}, \lambda_{\mathrm{ULD}}$.
    \Statex \textit{\# Preprocessing (cached across epochs)}
    \State $\mathbf{s}\gets\mathcal{T}_S(x),\quad \mathbf{t}\gets\mathcal{T}_T(x)$
    \State $\{(A_k^S,A_k^T)\}_{k=1}^{K}\gets\texttt{DPAlign}(\mathbf{s},\mathbf{t})$
    \Statex \textit{\# Forward passes}
    \State Run $f_S(\mathbf{s})$ with gradient and $f_T(\mathbf{t})$ without gradient.
    \Statex \textit{\# Per-chunk KD loss}
    \For{$k = 1, \ldots, K$}
      \State Merge chunk distributions $\hat p_S^{(k)}, \hat p_T^{(k)}$ via the inherited chain-rule merge.
      \If{$\mathcal{M} = \textsf{P-KL}$} \hfill \textit{(partition-free direct projection KL)}
        \State \rev{Project $W^{\!\top}\hat p_S^{(k)}$, restrict to retained teacher support, and normalize to obtain $\tilde p_S^{(k)}$.}
        \State $\mathcal{L}^{(k)} \gets \mathrm{KL}\bigl(\hat p_T^{(k)} \,\|\, \tilde p_S^{(k)}\bigr)$.
      \Else \hfill \textit{(H-KL: hybrid common-KL $+$ ULD on relaxed $\mathcal{C}$)}
        \State $\mathcal{L}^{(k)} \gets \lambda_{\mathrm{KL}}\mathcal{L}_{\mathrm{common}}^{(k)} + \lambda_{\mathrm{ULD}}\mathcal{L}_{\mathrm{ULD}}^{(k)}$.
      \EndIf
    \EndFor
    \Statex \textit{\# Loss aggregation and dynamic KD/CE scaling}
    \State $\mathcal{L}_{\mathrm{KD}} \gets \tau^2\cdot \tfrac{1}{K}\sum_{k=1}^{K}\mathcal{L}^{(k)}$.
    \State $\mathcal{L}_{\mathrm{CE}} \gets$ next-token cross-entropy of $f_S$ on $\mathbf{s}$.
    \State $\gamma \gets \mathrm{sg}\!\bigl(\mathcal{L}_{\mathrm{CE}} / \mathcal{L}_{\mathrm{KD}}\bigr)$ \hfill \textit{(stop-gradient)}
    \State $\mathcal{L} \gets \gamma\cdot\mathcal{L}_{\mathrm{KD}} + \mathcal{L}_{\mathrm{CE}}$.
    \State \rev{Update $f_S$; for learnable P-KL, also update the retained values of $W$ while fixing its topology.}
  \end{algorithmic}
\end{algorithm}

%% file: tex/experiment.tex
\section{Experiments}
\label{sec:experiments}

\paragraph{Teachers and students.}
\rev{We distill Llama-3.2-1B, Gemma-3-1B-pt, and OLMo-2-1B students from Qwen3-4B-Base and Phi-4-mini-Instruct. Llama-3.2-3B provides a same-tokenizer reference for Llama students.}

\paragraph{Student and training setup.}
For longer-training and multi-teacher experiments, we use \textbf{Llama-3.2-1B} and train on the Nemotron-ClimbMix dataset~\citep{diao2025nemotron} for $30{,}000$ steps with a batch size of $768$ and context length $4096$. We use AdamW with initial learning rate $5\!\times\!10^{-5}$, $5\%$ warmup followed by cosine decay to $0$, weight decay $0.1$, and gradient clipping at $1.0$. Distillation and cross-entropy losses are combined via Eq.~\eqref{eq:total_loss} with temperature $\tau{=}1.0$. The projection matrix $W$ is initialized from tokenizer-level string rules, truncated to the top-$4$ teacher tokens per student token, and jointly refined with the student under P-KL (LR $=10^{-2}$, no gradient clipping), while kept fixed under H-KL. For both GOLD and H-KL, we use $\lambda_{\mathrm{KL}}=\lambda_{\mathrm{ULD}}=1$.

\rev{Cross-tokenizer comparisons use 5,000 steps and context length 2,048 with the same batch size and optimization settings. Checkpoints, data, and evaluation are matched within each teacher--student setting.}

\paragraph{Evaluation.}
We evaluate 3-shot accuracy on five benchmarks spanning knowledge, mathematical reasoning, and commonsense: MMLU~\citep{hendrycks2020measuring},
GSM8k~\citep{cobbe2021training},
MATH-Hendrycks~\citep{hendrycks2021measuring},
Winogrande~\citep{sakaguchi2021winogrande}, and
HellaSwag~\citep{zellers2019hellaswag}.  All numbers are
reported on the official test splits using identical evaluation
settings across methods.

\begin{revision}
In the main comparisons, Avg. denotes the unweighted mean accuracy across these five benchmarks, and gains are reported in percentage points. Comparisons are matched within each training budget. We use lm-evaluation-harness v0.4.9.1~\citep{gao2021framework}. GSM8K and seven-category Hendrycks MATH use vanilla 3-shot prompts, greedy decoding, and a 256-token limit. GSM8K uses normalized numeric exact match; MATH extracts the final boxed reference and applies Hendrycks LaTeX normalization to the extracted prediction. Section~\ref{sec:cot_check} evaluates CoT prompting. These controlled base-model evaluations differ from public instruction-tuned scores.
\end{revision}

\paragraph{Baselines.}
\rev{We compare with \textbf{ULD}~\citep{boizard2024towards}, \textbf{GOLD}~\citep{patiño2025_unlocking_on_policy_distillation_for_any_model_family}, \textbf{ALM}~\citep{minixhofer2025universal}, \textbf{CDM}~\citep{chen2025cdm}, and \textbf{SeDi}~\citep{wang2026sedi}, alongside continued pre-training and same-tokenizer KD. ALM follows TokenKit; CDM and SeDi follow the authors' SeDi repository. All methods run in a common training stack, with matched hardware, precision, and optimization within each comparison. X-Token uses P-KL when task-critical tokens are unmatched and H-KL when their common-set coverage is reliable.}

\paragraph{Computational Resources.}
\rev{Runs are feasible on a single H100 with gradient accumulation; an allocation of 128 H100s supports concurrent experiments. The distributed profiling in Table~\ref{tab:runtime} measures end-to-end throughput.}

\input{tex/comparisons.tex}

\subsection{\rev{Longer Training and Multi-Teacher Distillation}}
\label{sec:main_results}

\begin{table*}[t]
  \caption{\rev{Longer-training and multi-teacher results on Llama-3.2-1B: 30,000 steps, context 4,096, mean accuracy across the five benchmarks (3-shot).}
    \textbf{Llama-1B}: Llama-3.2-1B; \textbf{Llama-3B}: Llama-3.2-3B; \textbf{Qwen-4B}: Qwen3-4B\rev{-Base}; \textbf{Phi-mini}: Phi-4-mini-Instruct.
    Teacher rows (\textit{italic}) report standalone performance and are excluded from best-in-column comparisons.
    All methods share identical settings except for the distillation loss and teacher configuration.
    \textbf{Bold} denotes the best student result per column.}  
  \label{tab:main}
  \centering
  \footnotesize
  \setlength{\tabcolsep}{3pt}
  \resizebox{0.9\linewidth}{!}{%
  \begin{tabular}{llcccccc}
    \toprule
    \textbf{Setting} & \textbf{Method}
      & \textbf{MMLU} & \textbf{GSM8k} & \textbf{MATH}
      & \textbf{WG} & \textbf{HS} & \textbf{Avg.} \\
    \midrule
    \multirow{2}{*}{No distillation}
      & Llama-1B (base)                        & 32.05 & 5.69  & 5.48 & 61.48 & \textbf{65.08} & 33.96 \\
      & Continued pre-training                 & 40.50 & 10.25 & 6.90 & 61.60 & 63.90 & 36.63 \\
    \midrule
    \multirow{3}{*}{\makecell[l]{Teachers\\(reference)}}
      & \textit{Llama-3B}                  & \textit{55.97} & \textit{24.94} & \textit{8.82}  & \textit{70.01} & \textit{74.92} & \textit{46.93} \\
      & \textit{Phi-mini}           & \textit{68.32} & \textit{82.71} & \rev{\textit{19.22}} & \textit{74.90} & \textit{73.36} & \rev{\textit{63.70}} \\
      & \textit{Qwen-4B}                      & \textit{72.43} & \textit{84.61} & \textit{27.76} & \textit{72.30} & \textit{75.00} & \textit{66.42} \\
    \midrule
    Same tokenizer
      & Llama-3B $\to$ 1B                       & 43.83 & 12.89 & 8.16 & 62.70 & 64.42 & 38.40 \\
    \midrule
    \multirow{6}{*}{\makecell[l]{Cross tokenizer\\(single teacher)}}
      & Qwen-4B, ULD                            & 40.34 & 14.56 & 4.04 & 61.96 & 62.93 & 36.77 \\
      & Qwen-4B, GOLD                           & 42.56 & 2.56  & 4.50 & 62.95 & 62.59 & 35.03 \\
      & Qwen-4B, \textbf{X-Token (P-KL)}        & 44.67 & 15.54 & 7.96 & 63.46 & 62.63 & \underline{38.85} \\
      & Phi-mini, ULD                           & 41.43 & 17.97 & 6.24 & 62.59 & 63.32 & 38.31 \\
      & Phi-mini, GOLD                          & 43.50 & 16.50 & 7.80 & 62.60 & 62.92 & 38.66 \\
      & Phi-mini, \textbf{X-Token (H-KL)}       & 43.93 & 19.11 & 8.32 & 61.87 & 62.67 & \underline{39.18} \\
    \midrule
    \multirow{3}{*}{Multi-teacher}
      & Phi-mini $+$ Llama-3B (X-Token)         & \textbf{46.32} & \textbf{20.39} & \textbf{9.02} & 63.3 & 63.38 & \textbf{40.48} \\
      & Phi-mini $+$ Qwen-4B (X-Token)         & 43.98 & 14.63 & 8.10 & 62.74 & 63.00 & 38.49 \\
      & Phi-mini $+$ Qwen-4B $+$ Llama-3B (X-Token) & 45.86 & 19.18 & 8.56 & \textbf{63.61} & 63.55 & 40.15 \\
    \bottomrule
  \end{tabular}%
  }
  \vspace{-0.1in}
\end{table*}

\rev{Table~\ref{tab:main} evaluates longer training and multi-teacher configurations under a separate, matched budget.} Continued pre-training of Llama-1B without a teacher yields only modest gains over the frozen baseline ($33.96 \!\to\! 36.63$ avg.), \rev{and remains below X-Token and same-tokenizer KD; GOLD on the Qwen pair is an exception, scoring below continued pre-training.} Same-tokenizer KD from Llama-3B reaches $38.40$ avg., providing a same-family reference for cross-tokenizer methods.
\paragraph{P-KL on Qwen-4B (uncommon-token regime).}
On the Qwen pair, multi-digit numerals fall into the uncommon subset, where GOLD's gradients suppress them. GOLD achieves $35.03$ avg.\ ($2.56$ on GSM8k), underperforming even continued pre-training without a teacher ($36.63$ avg.), indicating that its partition is harmful in this regime.

P-KL removes the partition and routes student mass through $W$ over teacher multi-token decompositions, improving to $38.85$ avg.\ ($+3.82$ over GOLD; $6.07{\times}$ on GSM8k, $2.56\!\to\!15.54$). This also surpasses same-tokenizer KD from Llama-3.2-3B ($12.89$ on GSM8k), showing that cross-tokenizer KD with P-KL can exceed same-family KD on math.

Since X-Token and GOLD share alignment and training setup, this gap isolates the loss formulation. Notably, pure ULD already improves over GOLD ($36.77$ vs.\ $35.03$), indicating that the partition is the primary source of failure. P-KL further outperforms ULD by $+2.08$ avg.\ by adding identity-aware projection.

\paragraph{H-KL on Phi-mini (sound-partition regime).}
On the Phi-mini pair, multi-digit numerals remain in the common subset, so GOLD's partition is structurally sound and achieves $38.66$ avg. H-KL relaxes strict string matching to top-ranked teacher mappings under $W$, improving coverage while retaining the hybrid loss. This yields $39.18$ avg.\ ($+0.52$ over GOLD), isolating the benefit of relaxed matching.

Pure ULD performs slightly worse ($38.31$, $-0.35$ vs.\ GOLD), consistent with the partition being well-formed: when critical tokens lie in the common set, dropping the partition and using P-KL sacrifices identity-aligned signal and leads to degradations as shown in Table~\ref{tab:loss_variant}.

\paragraph{Multi-teacher distillation gives complementary gains.}
Combining Phi-mini (H-KL) with Llama-3B (same-tokenizer) under static weighting reaches $40.48$ avg., exceeding the same-family reference by $+2.1$ and the best single-teacher cross-tokenizer run ($39.18$) by $+1.3$, demonstrating strong complementarity: Phi-mini contributes math/reasoning, while Llama-3B contributes commonsense knowledge.

In contrast, combining two cross-tokenizer reasoning teachers (Phi-mini $+$ Qwen-4B) achieves $38.49$, below the best single-teacher result, suggesting overlapping capabilities and interference. Adding Qwen-4B as a third teacher yields $40.15$ avg., similar overall but with trade-offs: math/reasoning degrades (MMLU $46.32\!\to\!45.86$, GSM8k $20.39\!\to\!19.18$, MATH $9.02\!\to\!8.56$) while commonsense improves slightly, again indicating redundancy or interference. \rev{These results favor complementary teacher strengths but do not establish an automatic selection rule.}

\input{tex/robustness.tex}

\subsection{Ablations and Design Checks}
\label{sec:ablations}

\rev{We ablate four design choices, with initialization, sparsity, and seed analyses in Appendix~\ref{app:projection_ablations}:} the loss mode (P-KL vs.\ H-KL), the projection matrix $W$ (frozen vs.\ learned), the teacher weighting strategy, and the KD/CE scaling scheme.

\begin{tcolorbox}[colback=gray!5!white,colframe=gray!75!black]
\begin{itemize}[leftmargin=4mm]
    \item \textbf{Finding 1:} Tokenizer-dependent loss selection is crucial for effective cross-tokenizer KD.

    \item \textbf{Finding 2:} P-KL is preferred when critical tokens fall outside the common set.

    \item \textbf{Finding 3:} H-KL performs better when token alignment is preserved, leveraging sharper identity-aligned supervision.

    \item \textbf{Finding 4:} Complementary teachers drive multi-teacher gains.

    \item \textbf{Finding 5:} Simple static weighting suffices for combining teacher signals.
\end{itemize}
\end{tcolorbox}

\paragraph{P-KL vs.\ H-KL on each teacher.}

This ablation shows that neither loss mode dominates: each exhibits a sharp drop when applied to the wrong teacher. We evaluate both P-KL and H-KL on each teacher with all else fixed (Table~\ref{tab:loss_variant}). P-KL outperforms H-KL by $+3.55$ avg.\ on Qwen3-4B\rev{-Base}, while H-KL outperforms P-KL by $+1.68$ on Phi-4-mini. This reversal aligns with the mechanism: P-KL partition-free projection is preferred when critical tokens fall in the uncommon set, whereas H-KL is favored when the partition is structurally sound. These results validate the per-teacher loss selection used in Table~\ref{tab:main}.

\begingroup
\setlength{\intextsep}{8pt}
\begin{table}[!htbp]
  \captionsetup{font=small,skip=5pt}
  \caption{Average accuracy across the five benchmarks for each
    loss mode on each teacher.  The two modes flip rankings between
    teachers (\textbf{bold}: per-teacher winner).}
  \label{tab:loss_variant}
  \centering
  \footnotesize
  \setlength{\tabcolsep}{4pt}
  \begin{tabular}{lcc}
    \toprule
    \textbf{Teacher} & \textbf{P-KL} & \textbf{H-KL} \\
    \midrule
    Qwen-4B    & \textbf{38.85} & 35.30 \\
    Phi-mini   & 37.50          & \textbf{39.18} \\
    \bottomrule
  \end{tabular}
\end{table}
\endgroup

\paragraph{Frozen vs.\ learnable projection matrix.}
Table~\ref{tab:ablation_proj} compares a frozen $W$ against jointly learning $W$ on the Qwen3-4B\rev{-Base} (P-KL) pair. Learning $W$ yields a modest but consistent improvement ($38.85$ vs.\ $38.37$ avg., improving four of the five benchmarks), indicating that the rule-based construction provides a strong initialization that can be further refined by the distillation loss with minimal overhead.

\begin{table}[t]
  \caption{Frozen vs.\ jointly learned projection matrix $W$ on Qwen3-4B\rev{-Base} (P-KL) with a Llama-3.2-1B student (3-shot). \rev{Joint learning is the P-KL default.} H-KL is omitted since $W$ affects it only via a discrete top-1 selection and receives no gradient. \textbf{Bold} indicates the best result per column.}
  \label{tab:ablation_proj}
  \centering
  \footnotesize
  \setlength{\tabcolsep}{3pt}
  \resizebox{\linewidth}{!}{%
  \begin{tabular}{lcccccc}
    \toprule
    \textbf{Projection matrix}
      & \textbf{MMLU} & \textbf{GSM8k} & \textbf{MATH}
      & \textbf{WG} & \textbf{HS} & \textbf{Avg.} \\
    \midrule
    \rev{Frozen}
                                     & 43.36 & \textbf{15.77} & 7.94
                                     & 62.59 & 62.17 & 38.37 \\
    \rev{Learned (default)}
                                     & \textbf{44.67} & 15.54 & \textbf{7.96}
                                     & \textbf{63.46} & \textbf{62.63} & \textbf{38.85} \\
    \bottomrule
  \end{tabular}
  }
\end{table}

\paragraph{Dynamic KD/CE scaling.}

\begin{table}[t]
  \caption{Dynamic KD/CE scaling vs.\ fixed-weight combinations
    on the Llama-3.2-1B student with the Qwen3-4B\rev{-Base} teacher (P-KL),
    $3{,}000$ training steps, 3-shot. \rev{Fixed weights are $(\lambda_{\mathrm{KL}},\lambda_{\mathrm{CE}})$.} \textbf{Bold}: best per
    column.}
  \label{tab:ablation_dls}
  \centering
  \footnotesize
  \setlength{\tabcolsep}{4pt}
  \resizebox{\linewidth}{!}{%
  \begin{tabular}{lcccccc}
    \toprule
    \textbf{KD/CE combination}
      & \textbf{MMLU} & \textbf{GSM8k} & \textbf{MATH}
      & \textbf{WG}   & \textbf{HS}    & \textbf{Avg.} \\
    \midrule
    \rev{Fixed $(1.0,0.1)$}
                                & 40.08 & 8.49 & 5.88 & 62.19 & 62.97 & 35.92 \\
    \rev{Fixed $(0.5,0.5)$}
                                & 40.07 & 9.48 & 6.00 & 62.59 & 63.22 & 36.27 \\
    \rev{Fixed $(0.1,1.0)$}
                                & 39.75 & 8.57 & 5.98 & \textbf{63.14} & \textbf{63.81} & 36.25 \\
    \textbf{Dynamic (default)}  & \textbf{40.15} & \textbf{9.70} & \textbf{6.04} & \textbf{63.14} & 62.94 & \textbf{36.39} \\
    \bottomrule
  \end{tabular}
  }
\end{table}

Dynamic scaling rebalances KD and CE at each step based on
their relative magnitudes (Eq.~\eqref{eq:total_loss}).
Table~\ref{tab:ablation_dls} compares it to three
fixed-weight settings spanning KD-heavy
($\lambda_{\mathrm{KL}}{=}1.0,\lambda_{\mathrm{CE}}{=}0.1$),
balanced ($0.5/0.5$), and CE-heavy ($0.1/1.0$) regimes on the
Qwen3-4B\rev{-Base} (P-KL) teacher. All configurations are run for $3{,}000$ steps
to keep the four-way sweep tractable. \rev{Dynamic scaling improves over the strongest fixed alternative by 0.12 points while avoiding per-setting coefficient tuning.}

\begin{table*}[t]
  \centering
  \footnotesize
  \setlength{\tabcolsep}{3pt}
  \begin{minipage}{0.65\linewidth}
    \centering
    \caption{Multi-teacher weighting on Phi-4-mini $+$
      Llama-3.2-3B with a Llama-3.2-1B student, $30{,}000$
      steps, 3-shot.  \textbf{Bold}: best per column among
      complete rows.  Static weights given as
      $(\alpha_{\mathrm{Llama}}, \alpha_{\mathrm{Phi}})$.}
    \label{tab:ablation_mt_static}
    \resizebox{\linewidth}{!}{%
    \begin{tabular}{lcccccc}
      \toprule
      \textbf{Weighting}
        & \textbf{MMLU} & \textbf{GSM8k} & \textbf{MATH}
        & \textbf{WG}   & \textbf{HS}    & \textbf{Avg.} \\
      \midrule
      Static $(0.8, 0.2)$
                            & 44.48 & 14.10 & 8.62 & 62.51 & \textbf{64.09} & 38.76 \\
      Static $(0.5, 0.5)$
                            & 45.97 & 19.56 & 8.82 & 63.14 & 63.98 & 40.29 \\
      \textbf{Static $(0.2, 0.8)$}
                            & \textbf{46.32} & \textbf{20.39} & \textbf{9.02} & 63.30 & 63.38 & \textbf{40.48} \\
      \midrule
      Adaptive (CE)         & 45.84 & 18.80 & 9.04 & \textbf{63.54} & 63.85 & 40.21 \\
      Adaptive (entropy)    & 45.63 & 18.35 & 8.54 & 62.90 & 63.65 & 39.81 \\
      Adaptive (max-prob)   & 45.61    & 19.11    & 8.82   & 63.38    & 63.88    & 40.16 \\
      \bottomrule
    \end{tabular}}
  \end{minipage}
  \hfill
  \begin{minipage}{0.3\linewidth}
    \centering
    \caption{Multi-teacher weighting on Phi-4-mini $+$ Qwen3-4B\rev{-Base}
      with a Llama-3.2-3B student, $30{,}000$ steps.  Static
      ratios given as $(\alpha_{\mathrm{Phi}}, \alpha_{\mathrm{Qwen}})$.}
    \label{tab:mt_weighting}
    \vspace{-0.1in}
    \resizebox{0.95\linewidth}{!}{%
    \begin{tabular}{lc}
      \toprule
      \textbf{Weighting} & \textbf{Avg.} \\
      \midrule
      Static $(0.5, 0.5)$  & 51.48 \\
      Adaptive (max-prob)  & 51.30 \\
      Adaptive (CE)        & 51.33 \\
      Adaptive (entropy)   & 51.59 \\
      \textbf{Static $(0.8, 0.2)$} & \textbf{52.19} \\
      \bottomrule
    \end{tabular}}
  \end{minipage}
\end{table*}

\paragraph{Multi-Teacher weighting.}
We compare static and confidence-adaptive softmax parameterizations of the per-teacher weight $\alpha_m$ (Eq.~\eqref{eq:multiteacher}) across two setups. Table~\ref{tab:ablation_mt_static} (1B student; Phi-4-mini $+$ Llama-3.2-3B) evaluates three static ratios and \rev{three} adaptive variants, while Table~\ref{tab:mt_weighting} (3B student; Phi-4-mini $+$ Qwen3-4B\rev{-Base}) reports analogous results.

Across both, Phi-heavy static weighting performs best: $(0.2,0.8)$ reaches $40.48$ avg.\ on the 1B run and $(0.8,0.2)$ reaches $52.19$ on the 3B run, both exceeding adaptive schemes (CE: $40.21/51.33$, entropy: $39.81/51.59$, max-prob: $40.16/51.30$). This supports our observation that adaptive weighting adds tuning complexity without consistent gains. We therefore adopt static weighting in the main results.

%% file: tex/comparisons.tex
\begin{revision}
\subsection{Comparison with Baseline Methods}
\label{sec:matched_results}
Table~\ref{tab:matched_baselines} compares all evaluated cross-tokenizer methods under a common training protocol. X-Token achieves the highest average accuracy in each teacher--student setting. P-KL exceeds the strongest competing baseline, CDM, by 0.60 percentage points on Qwen-to-Llama and ULD by 0.55 on Qwen-to-OLMo. H-KL exceeds GOLD by 0.04 points on Phi-to-Llama and 0.13 on Qwen-to-Gemma, where critical-token coverage is reliable.

\begin{table*}[t]
\begin{revision}
\centering
\footnotesize
\setlength{\tabcolsep}{1.5pt}
\renewcommand{\arraystretch}{1.05}
\captionsetup{skip=5pt}
\caption{Baseline comparisons at 5,000 steps, batch size 768, and context length 2,048. Avg. is the unweighted mean accuracy (\%) across MMLU, GSM8K, MATH, WinoGrande (WG), and HellaSwag (HS). Bold marks the best result within each teacher--student block. Only evaluated methods are shown.}
\label{tab:matched_baselines}
\begin{minipage}[t]{0.485\textwidth}
\vspace{0pt}
\begin{tabular*}{\linewidth}{@{}l@{\extracolsep{\fill}}rrrrrr@{}}
\toprule
Method & MMLU & GSM8K & MATH & WG & HS & Avg. \\
\midrule
\multicolumn{7}{l}{\textbf{Qwen3-4B-Base $\to$ Llama-3.2-1B}} \\
\addlinespace[2pt]
ULD & 33.09 & 9.40 & 5.34 & 62.51 & \textbf{63.52}  & 34.77 \\
GOLD & \textbf{41.11} & 1.97 & 4.00 & 62.98 & 62.65  & 34.54 \\
ALM & 34.49 & 8.57 & 5.72 & 61.88 & 62.47  & 34.63 \\
CDM & 40.70 & 9.70 & 5.58 & 62.59 & 62.27  & 36.17 \\
SeDi & 37.18 & 9.55 & 5.82 & 62.75 & 60.54  & 35.17 \\
X-Token (P-KL) & 41.09 & \textbf{10.46} & \textbf{5.88} & \textbf{63.46} & 62.95  & \textbf{36.77} \\
\midrule
\multicolumn{7}{l}{\textbf{Phi-4-mini-Instruct $\to$ Llama-3.2-1B}} \\
\addlinespace[2pt]
ULD & 38.53 & 10.54 & 5.92 & \textbf{63.85} & \textbf{63.84}  & 36.54 \\
GOLD & 42.08 & 12.81 & \textbf{7.28} & 62.90 & 63.25  & 37.66 \\
ALM & 36.50 & 11.68 & 5.80 & 62.27 & 62.98  & 35.85 \\
CDM & 40.96 & 11.22 & 6.32 & 62.35 & 62.42  & 36.65 \\
SeDi & 38.16 & 9.25 & 5.52 & 62.59 & 60.57  & 35.22 \\
X-Token (H-KL) & \textbf{42.12} & \textbf{12.89} & 7.22 & 63.06 & 63.22  & \textbf{37.70} \\
\bottomrule
\end{tabular*}
\end{minipage}
\hfill
\begin{minipage}[t]{0.485\textwidth}
\vspace{0pt}
\begin{tabular*}{\linewidth}{@{}l@{\extracolsep{\fill}}rrrrrr@{}}
\toprule
Method & MMLU & GSM8K & MATH & WG & HS & Avg. \\
\midrule
\multicolumn{7}{l}{\textbf{Qwen3-4B-Base $\to$ Gemma-3-1B-pt}} \\
\addlinespace[2pt]
ULD & 26.78 & 1.29 & 3.18 & 61.88 & \textbf{61.18}  & 30.86 \\
GOLD & 33.59 & 6.22 & 3.78 & 61.64 & 59.80  & 33.01 \\
ALM & 25.78 & 2.73 & 3.34 & 61.71 & 60.33  & 30.78 \\
SeDi & 28.00 & 4.62 & 3.42 & 59.43 & 59.33  & 30.96 \\
X-Token (P-KL) & 32.05 & 6.22 & \textbf{3.84} & \textbf{62.51} & 60.89  & 33.10 \\
X-Token (H-KL) & \textbf{34.33} & \textbf{6.52} & 3.80 & 61.25 & 59.77  & \textbf{33.13} \\
\midrule
\multicolumn{7}{l}{\textbf{Qwen3-4B-Base $\to$ OLMo-2-1B}} \\
\addlinespace[2pt]
ULD & 43.01 & 31.61 & 5.04 & \textbf{66.93} & \textbf{65.68}  & 42.45 \\
GOLD & 44.88 & 4.09 & 3.90 & 65.75 & 65.24  & 36.77 \\
X-Token (P-KL) & \textbf{44.92} & \textbf{32.83} & \textbf{5.72} & 65.98 & 65.57  & \textbf{43.00} \\
X-Token (H-KL) & 44.84 & 3.79 & 3.92 & 65.90 & 65.27  & 36.74 \\
\bottomrule
\end{tabular*}
\end{minipage}
\end{revision}
\end{table*}

OLMo reproduces the uncommon-token failure: GOLD and H-KL reach only 4.09 and 3.79 on GSM8K, while P-KL reaches 32.83 and improves average accuracy over GOLD by 6.23 percentage points. On Gemma, where critical-token coverage is reliable, H-KL performs best. Together, these results support coverage-guided loss selection across student families.
\end{revision}

%% file: tex/robustness.tex
\begin{revision}
\subsection{Robustness to Math Evaluation Protocol}
\label{sec:cot_check}
We re-evaluate Qwen-to-Llama students with 8-shot CoT on GSM8K and 4-shot CoT on Hendrycks MATH, each with a 1,024-token limit (Table~\ref{tab:cot}). P-KL retains the highest mean math accuracy and overall average accuracy, although individual task rankings change: SeDi leads on MATH. The overall advantage therefore persists under stronger reasoning prompts and longer generation.

\begin{table*}[t]
\begin{revision}
\centering
\small
\setlength{\tabcolsep}{4pt}
\caption{CoT re-evaluation of the Qwen3-4B-Base $\to$ Llama-3.2-1B students from Table~\ref{tab:matched_baselines}. Vanilla: 3-shot, 256 tokens on both math tasks. CoT: 8-shot GSM8K and 4-shot MATH, each with a 1,024-token limit. The updated five-task average replaces only the two math scores; the other three scores are unchanged. Bold marks the best value in each column.}
\label{tab:cot}
\begin{tabular}{lrrrrrrrr}
\toprule
 & \multicolumn{2}{c}{GSM8K} & \multicolumn{2}{c}{MATH} & \multicolumn{2}{c}{Math avg.} & \multicolumn{2}{c}{Five-task avg.} \\
\cmidrule(lr){2-3}\cmidrule(lr){4-5}\cmidrule(lr){6-7}\cmidrule(lr){8-9}
Method & Vanilla & CoT & Vanilla & CoT & Vanilla & CoT & Vanilla & CoT \\
\midrule
ULD & 9.40 & 12.28 & 5.34 & 6.10 & 7.37 & 9.19  & 34.77 & 35.50 \\
GOLD & 1.97 & 2.88 & 4.00 & 4.08 & 2.99 & 3.48  & 34.54 & 34.74 \\
ALM & 8.57 & 10.77 & 5.72 & 4.52 & 7.15 & 7.65  & 34.63 & 34.83 \\
CDM & 9.70 & 14.03 & 5.58 & 7.32 & 7.64 & 10.68  & 36.17 & 37.38 \\
SeDi & 9.55 & 14.71 & 5.82 & \textbf{7.42} & 7.69 & 11.07  & 35.17 & 36.52 \\
X-Token (P-KL) & \textbf{10.46} & \textbf{15.16} & \textbf{5.88} & 7.28 & \textbf{8.17} & \textbf{11.22}  & \textbf{36.77} & \textbf{37.99} \\
\bottomrule
\end{tabular}
\end{revision}
\end{table*}

\subsection{Runtime and Memory}
\label{sec:runtime}
For Llama's 128,256-token vocabulary, top-$4$ storage occupies 4.89 MB, with one-time construction taking $57.79\pm0.32$ s. Each teacher requires one cached projection. DP alignment takes 134.3 ms/sample at 2K tokens and 306.7 ms at 4K; CPU prefetching overlaps it with GPU computation, with no observed steady-state delay. ULD, GOLD, and X-Token share these alignments.

Table~\ref{tab:runtime} reports end-to-end step times. P-KL and H-KL take 0.747 and 0.885 s/step, respectively, and are faster than all evaluated cross-tokenizer baselines. P-KL adds 26.6\% over the standard-KD timing reference.

\begin{table}[t]
\begin{revision}
\centering
\small
\setlength{\tabcolsep}{4pt}
\caption{End-to-end profiling at batch size 768 and context 2,048, averaged over steps 10--99. Runs share the teacher and distributed configuration. Standard KD omits cross-tokenizer operations; overhead is relative to its 0.590 s/step.}
\label{tab:runtime}
\begin{tabular}{lrr}
\toprule
Method & Seconds/step & Overhead \\
\midrule
Standard KD reference & 0.590 & -- \\
X-Token P-KL & 0.747 & 26.6\% \\
X-Token H-KL & 0.885 & 50.0\% \\
GOLD & 0.927 & 57.1\% \\
ALM & 1.170 & 98.3\% \\
ULD & 1.579 & 167.6\% \\
SeDi & 1.694 & 187.1\% \\
CDM & 5.252 & 790.2\% \\
\bottomrule
\end{tabular}
\end{revision}
\end{table}

\paragraph{Implementation.}
Projection uses PyTorch sparse-CSR \texttt{torch.matmul} followed by teacher-support selection, without a custom kernel. Learning $W$ adds 0.012 s/step and is included in P-KL timing. All methods share the pipeline, hardware, precision, and optimization stack. P-KL avoids full-vocabulary sorting and dynamic matching; the measured ordering remains implementation-dependent.
\end{revision}

%% file: tex/related.tex
\section{Related Work}
\label{sec:related}

\begin{revision}
Cross-tokenizer transfer aligns output distributions or learned representations, or adapts the model vocabulary. X-Token preserves the student tokenizer and inference architecture.
\end{revision}

\subsection{Logit-distribution-based methods}
\textbf{ULD}~\citep{boizard2024towards} sidesteps vocabulary
alignment by rank-sorting both distributions and minimizing an
$L_1$ distance, discarding token identity.
\textbf{GOLD}~\citep{patiño2025_unlocking_on_policy_distillation_for_any_model_family} adds span alignment, chain-rule chunk
aggregation, and a hybrid loss that partitions tokens into a
1-to-1 string-equality common set (direct KL) and an uncommon
remainder (ULD on the tail); \rev{it provides the partition-based reference for our failure analysis.}
\textbf{ALM}~\citep{minixhofer2025universal} aligns student and
teacher at the byte level, aggregates chunk-level log-probabilities,
and applies a Binary Cross Entropy/KL-style loss.
\textbf{MinED}~\citep{wan2024knowledge} maps each student token
to the closest teacher token under string edit distance, yielding
a rule-based 1-to-1 alignment baseline.

\begin{revision}
\textbf{CDM}~\citep{chen2025cdm} uses context to improve sequence alignment and dynamically map vocabularies. \textbf{MultiLevelOT}~\citep{cui2025multilevel} aligns teacher and student logit distributions at token and sequence levels using optimal transport with multiple cost matrices. \textbf{SeDi}~\citep{wang2026sedi} combines bipartite tokenization alignment with sliding-window re-encoding into the student's vocabulary, and further matches distributional information through entropy-aware transfer. These approaches make vocabulary correspondence or distribution matching explicit; X-Token instead reuses a tokenizer-initialized sparse one-to-many operator and distinguishes harmful-partition and reliable-coverage regimes with P-KL and H-KL.

\end{revision}

\subsection{Representation-based methods}
This family aligns teacher and student at the embedding or hidden-state level, typically requiring auxiliary trainable components or architectural modifications.
\textbf{DSKD}~\citep{zhang2024dual} \rev{uses learned projections and cross-model alignment to distill across heterogeneous representation and output spaces.}
\textbf{ZETT}~\citep{minixhofer2024zero} generates embeddings for a new vocabulary using a hypernetwork conditioned on token strings, enabling tokenizer transfer at the embedding level; in practice, the ALM pipeline~\citep{minixhofer2025universal} combines this with a logit-level loss.
\rev{\textbf{GenRecal}~\citep{lee2025genrecal} aligns heterogeneous vision-language models through a learned feature recalibrator, offering a representation-level approach to cross-model transfer.}
\textbf{BLD}~\citep{singh2026cross} converts teacher token distributions to byte-level distributions and augments the student with auxiliary byte-projection heads (discarded at inference), modifying the architecture to handle low-overlap vocabularies.

\begin{revision}
\subsection{Vocabulary adaptation}
\textbf{TokAlign}~\citep{li2025tokalign} aligns token IDs, replaces the vocabulary, and fine-tunes the adapted model. Its vocabulary-adaptation stage differs from X-Token's fixed-tokenizer setting.
\end{revision}

%% file: tex/appendix.tex
\section{Suppressive Gradients From The Common-KL Term}
\label{app:suppression_proof}

GOLD's common-KL term $\mathcal{L}_{\mathrm{common}}$ (Eq.~\ref{eq:gold_common}) is a
sum over matched columns of full-vocab softmaxes; the dependency
on $\log Z_{\mathrm{full}}$ inside $\log p_S[i]$ propagates
gradient back to every uncommon student logit, even though those
logits do not appear in the loss.

\begin{proposition}[Common-KL induces a suppressive gradient on uncommon logits]
  \label{prop:suppression}
  Let $z\in\mathbb{R}^{|\mathcal{V}_S|}$ be the student logits in
  a chunk, $p_S = \mathrm{softmax}(z)$, and let $p_T$ be the
  (fixed) teacher distribution.  Let $\mathcal{C}_T$ be the
  teacher-side projection of the common subset and
  $\mathcal{U}=\mathcal{V}_S\setminus\mathcal{C}_S$ the uncommon
  set.  Then for every uncommon student logit $j\in\mathcal{U}$,
  \begin{equation}
  \begin{split}
    \frac{\partial \mathcal{L}_{\mathrm{common}}}{\partial z_j}
    &= p_S[j] \cdot M_{\mathcal{C}}(T) \geq 0, \\
    \text{where} \quad M_{\mathcal{C}}(T) &:= \sum_{t \in \mathcal{C}_T} p_T[t] \in [0, 1].
  \end{split}
  \label{eq:suppression_gradient}
\end{equation}
  Because the gradient is non-negative, gradient descent with step
  $\eta > 0$ decreases $z_j$ at every step:
  $\Delta z_j = -\eta\,(\partial\mathcal{L}_{\mathrm{common}}/\partial z_j)
  = -\eta\,p_S[j]\,M_{\mathcal{C}}(T) \leq 0$.
  Since the softmax is monotonically increasing in each logit,
  driving $z_j$ downward shrinks $p_S[j]$ relative to all other
  student probabilities --- the probability mass of every uncommon
  token is suppressed, even though no uncommon token appears in
  $\mathcal{L}_{\mathrm{common}}$ and independent of the
  ground-truth token at the position.
\end{proposition}

\paragraph{Setup.}
We fix a chunk and let
$z\in\mathbb{R}^{|\mathcal{V}_S|}$ be the student logits.  The
full-vocab softmax is
\begin{equation}
  p_S[s] \;=\; \frac{\exp(z_s)}{Z_{\mathrm{full}}},
  \;\;
  Z_{\mathrm{full}} \;=\; \sum_{s'\in\mathcal{V}_S}\exp(z_{s'}),
  \label{eq:softmax_def}
\end{equation}
and $p_T$ is a fixed distribution over $\mathcal{V}_T$ that does
not depend on $z$.  The bijective common subset
$\mathcal{C}\subseteq\mathcal{V}_S\times\mathcal{V}_T$ has
projections $\mathcal{C}_S$ and $\mathcal{C}_T$, and we write
$\mathcal{U}=\mathcal{V}_S\setminus\mathcal{C}_S$.  The common-KL
term (Eq.~\ref{eq:gold_common}) is
\begin{equation}
  \mathcal{L}_{\mathrm{common}}(z)
  \;=\; \sum_{(s,t)\in\mathcal{C}}
        p_T[t]\,\bigl(\log p_T[t] - \log p_S[s]\bigr).
  \label{eq:common_kl_appendix}
\end{equation}

\paragraph{Preliminary identities.}
Treating distinct logits as independent variables, for any
$s,j\in\mathcal{V}_S$,
\begin{equation}
  \frac{\partial z_s}{\partial z_j}
    = \mathbf{1}[s = j], \;\;
  \frac{\partial \log Z_{\mathrm{full}}}{\partial z_j}
    = \frac{\exp(z_j)}{Z_{\mathrm{full}}}
    = p_S[j].
  \label{eq:partials}
\end{equation}

The first identity is immediate.  The second follows from
$\log Z_{\mathrm{full}} = \log\sum_{s'}\exp(z_{s'})$ by the
chain rule.

Combining with $\log p_S[s] = z_s - \log Z_{\mathrm{full}}$,
\begin{equation}
  \frac{\partial \log p_S[s]}{\partial z_j}
    \;=\; \mathbf{1}[s=j] \;-\; p_S[j].
  \label{eq:logp_partial}
\end{equation}

\paragraph{Proof of Proposition~\ref{prop:suppression}.}
Fix $j\in\mathcal{U}$.  Since $\mathcal{C}_S$ and $\mathcal{U}$
are disjoint, every $s\in\mathcal{C}_S$ satisfies $s\neq j$, so
$\mathbf{1}[s=j]=0$ in Eq.~\eqref{eq:logp_partial} and thus
$\partial\log p_S[s]/\partial z_j = -p_S[j]$ for every
$s\in\mathcal{C}_S$.  The teacher term $p_T[t]\log p_T[t]$ in
Eq.~\eqref{eq:common_kl_appendix} has no dependence on $z$.
Differentiating Eq.~\eqref{eq:common_kl_appendix} with respect
to $z_j$ therefore yields
\begin{align}
  \frac{\partial \mathcal{L}_{\mathrm{common}}}{\partial z_j}
    &= -\sum_{(s,t)\in\mathcal{C}} p_T[t] \cdot \frac{\partial \log p_S[s]}{\partial z_j} \notag \\
    &= -\sum_{(s,t)\in\mathcal{C}} p_T[t] \cdot \bigl(-p_S[j]\bigr) \notag \\
    &= p_S[j] \sum_{t\in\mathcal{C}_T} p_T[t] = p_S[j] M_{\mathcal{C}}(T) \label{eq:gradient_derivation}
\end{align}
where the second-to-last equality uses the bijection between
$\mathcal{C}_S$ and $\mathcal{C}_T$ (each $t\in\mathcal{C}_T$
appears in exactly one pair $(s,t)\in\mathcal{C}$).
Since $p_S[j]\geq 0$ and $M_{\mathcal{C}}(T)\in[0,1]$ (the
teacher is a probability distribution, so
$\sum_{t\in\mathcal{V}_T}p_T[t]=1$ and
$\mathcal{C}_T\subseteq\mathcal{V}_T$), the gradient is
non-negative and vanishes only when one of the two factors is
zero.  Under gradient descent with step $\eta>0$,
$\Delta z_j = -\eta\,p_S[j]\,M_{\mathcal{C}}(T)\leq 0$.  No
quantity in the derivation depends on the ground-truth token,
establishing both claims of
Proposition~\ref{prop:suppression}: the gradient is
non-negative on every uncommon logit, and its dependence on
$p_T$ alone makes it independent of the ground-truth token at
the position.
\hfill$\square$

\section{Algorithm Details}
\label{app:algorithm}

\paragraph{DP span alignment scoring and recurrence.}
For each training sequence we precompute a set of \emph{aligned
chunks} $\{(A_k^S,A_k^T)\}_{k=1}^{K}$ via a dynamic-programming
span alignment, where each pair of spans decodes to the same text
\rev{substring. Alignment is cached, prefetched on CPU, and overlapped with GPU computation. Its measured cost is 134.3 ms/sample at 2K and 306.7 ms/sample at 4K; no additional exposed steady-state wall-clock time was observed in the profiled pipeline. ULD, GOLD, and X-Token use the same alignments.}  Let $D(i,j)$ denote the maximum score achievable over
student prefix $\mathbf{s}_{1:i}$ and teacher prefix
$\mathbf{t}_{1:j}$; the recurrence is:
\begin{equation}
\resizebox{0.98\linewidth}{!}{$\displaystyle
  D(i,j) = \max \!
  \begin{cases}
    D(i-1, j-1) + \mathrm{match}(s_i, t_j) \\ 
    \hfill \text{(diagonal, 1-to-1)} \\[5pt]
    \displaystyle \max_{2\leq k\leq L} D(i-1, j-k) + \alpha_{\mathrm{comb}} k \cdot \mathbb{1}[s_i \equiv \mathbf{t}_{j-k+1:j}] \\ 
    \hfill \text{(1-to-}k\text{ combination)} \\[5pt]
    \displaystyle \max_{2\leq k\leq L} D(i-k, j-1) + \alpha_{\mathrm{comb}} k \cdot \mathbb{1}[\mathbf{s}_{i-k+1:i} \equiv t_j] \\ 
    \hfill \text{(}k\text{-to-1 combination)} \\[5pt]
    D(i-1, j) + \alpha_{\mathrm{gap}} \hfill \text{(gap in teacher)} \\[2pt]
    D(i, j-1) + \alpha_{\mathrm{gap}} \hfill \text{(gap in student)}
  \end{cases}
$}
  \label{eq:dp_recurrence}
\end{equation}
where $\mathrm{match}(s_i,t_j) = +\alpha_{\mathrm{exact}}$ if the two
(canonicalized) tokens agree and $-\alpha_{\mathrm{exact}}$ otherwise, and
``$\equiv$'' denotes canonicalized string equality between a single token
and the concatenation of a span.  The boundary conditions are
$D(i, 0) = i\cdot\alpha_{\mathrm{gap}}$ and
$D(0, j) = j\cdot\alpha_{\mathrm{gap}}$.  We use:
\begin{equation}
  \alpha_{\mathrm{exact}} = 3,\quad
  \alpha_{\mathrm{comb}} = 1.5,\quad
  \alpha_{\mathrm{gap}} = -1.5,
\end{equation}
in all experiments.  A backtrace from $D(n, m)$ recovers the set of
aligned chunks; transitions selected as gaps produce token positions
that are marked unaligned and excluded from the loss.

\paragraph{Why soft scoring.}
A hard-constraint DP (align-or-fail) has two failure modes on realistic
data: (i)~a local tokenization edge case (a byte-fallback token, an
unusual whitespace glyph) makes the entire sequence misalign or propagates
error into neighbouring chunks; (ii)~two locally-plausible alignments
tie, and an arbitrary tie-breaker produces inconsistent alignments across
training runs.  The scoring formulation resolves both: gaps cost
$|\alpha_{\mathrm{gap}}|$, so the DP prefers to insert a single gap
rather than distort a long stretch, and mismatched diagonals are
dominated by gap sequences whenever two or more consecutive positions
would otherwise mismatch.  The score parameters were chosen so that
(a)~$\alpha_{\mathrm{exact}} > |\alpha_{\mathrm{gap}}|$ to reward
alignment over walking around it, and (b)~a $k$-token combination
($+\alpha_{\mathrm{comb}}k$) competes favourably with $k$ individual
1-to-1 matches ($+\alpha_{\mathrm{exact}}k$) when the exact span-level
match is available.  We did not tune these values per dataset.

\paragraph{Failure mode of TRL surface-substring alignment.}
An alternative to surface-level DP, used in TRL's%
\footnote{\url{https://github.com/huggingface/trl}}
\textsc{Gold} trainer, pairs tokens by substring equality on
incrementally-decoded text: per-side decoded buffers are extended one
piece at a time and an alignment group is flushed whenever the two
buffers compare equal as raw strings. The algorithm is brittle in a
specific way: any byte-level disagreement between the two decoded
streams that is not canceled by a later piece prevents future
flushes, and the end-of-sequence force-flush dumps everything from
the divergence point onward into a single mis-grouped bucket.
Table~\ref{tab:trl_failure} shows a routine setting where this
occurs in cross-tokenizer KD; DP recovers the alignment via a single
gap move.

\begin{table}[t]
\caption{Failure mode of TRL surface-substring alignment under
default-configuration BOS asymmetry. The Llama-3 tokenizer
auto-prepends \texttt{<bos>} (= \texttt{<|begin\_of\_text|>}) under
\texttt{add\_bos\_token=True} (its config default) while Qwen-3 and
Phi-4-mini-Instruct default to \texttt{False}. Same input string
\texttt{"Hello world."} on both sides; decoded streams differ on
byte~0. Blue cells are student tokens, orange cells are teacher
tokens. \textbf{TRL alignment} (top block) emits a single
\textsc{super-group} bundling all student and teacher tokens
together. \textbf{DP alignment} (bottom block) emits one
alignment pair per row: the spurious \texttt{<bos>} is marked
as a one-sided \textsc{gap}, and the remaining tokens are
diagonal \textsc{match}es.}
\label{tab:trl_failure}
\centering
\footnotesize
\setlength{\fboxsep}{1.5pt}
\setlength{\fboxrule}{0.4pt}
\setlength{\tabcolsep}{5pt}
\renewcommand{\arraystretch}{1.20}
\resizebox{\linewidth}{!}{%
\begin{tabular}{@{}l|l|l@{}}
\toprule
\textbf{Pair} & \textbf{Student tokens} & \textbf{Teacher tokens} \\
\midrule
\multicolumn{3}{@{}l}{\textit{Input.}} \\
\addlinespace[2pt]
        & \stok{<bos>}\,\stok{Hello}\,\stok{ world}\,\stok{.}
        & \ttok{Hello}\,\ttok{ world}\,\ttok{.} \\
\midrule
\multicolumn{3}{@{}l}{\textbf{TRL alignment}} \\
\addlinespace[2pt]
\rowcolor{red!8}
\textcolor{red!70!black}{\xmark\,\textsc{super-group}}\ \#1
        & $\bigl\{$\,\stok{<bos>}\,\stok{Hello}\,\stok{ world}\,\stok{.}\,$\bigr\}$
        & $\bigl\{$\,\ttok{Hello}\,\ttok{ world}\,\ttok{.}\,$\bigr\}$ \\
\midrule
\multicolumn{3}{@{}l}{\textbf{DP alignment}} \\
\addlinespace[2pt]
\textcolor{red!70!black}{\xmark\,\textsc{gap}}\ \#1   & \stok{<bos>}     & \textit{(no teacher token)}    \\
\textcolor{green!45!black}{\cmark\,\textsc{match}}\ \#2 & \stok{Hello}    & \ttok{Hello}                   \\
\textcolor{green!45!black}{\cmark\,\textsc{match}}\ \#3 & \stok{ world}   & \ttok{ world}                  \\
\textcolor{green!45!black}{\cmark\,\textsc{match}}\ \#4 & \stok{.}        & \ttok{.}                       \\
\bottomrule
\end{tabular}
}
\end{table}

\textbf{Why TRL fails.}
TRL accumulates per-side decoded buffers and only flushes when buffers
compare equal as raw strings. After the first piece is appended,
$s_\mathrm{buf}{=}$ \texttt{"<bos>"} (= \texttt{"<|begin\_of\_text|>"},
16 chars) vs.\ $t_\mathrm{buf}{=}$ \texttt{"Hello"} (5 chars).
Length-driven extension keeps the two buffers character-misaligned
through every prefix, so the buffer-equality flush never fires; the
end-of-sequence force-flush emits both sides as a single
\textsc{super-group} (Pair~\#1 in the top block of
Table~\ref{tab:trl_failure}).

\textbf{Why DP works.}
DP's recurrence has explicit gap moves at fixed cost. It marks the
spurious \texttt{<bos>} as a one-sided \textsc{gap} of unit cost
(Pair~\#1 in the bottom block) and aligns the three content tokens
diagonally as 1-to-1 \textsc{match}es (Pairs~\#2--\#4). The
disagreement is localized to a single \textsc{gap} pair regardless of
how long the sentence is.

\paragraph{Projection matrix as a probability-preserving operator.}
Because each row of $W$ is non-negative and sums to 1, left-multiplication
by $W^{\!\top}$ acts as a convex combination of rows and is
probability-preserving: if $\mathbf{p}_S$ is a probability vector, then
$\mathbf{W}^{\!\top}\mathbf{p}_S$ is also a probability vector over
$\mathcal{V}_T$.  In particular,
\begin{align}
  \sum_{t\in\mathcal{V}_T} \bigl(W^{\!\top}\mathbf{p}_S\bigr)[t]
    &= \sum_{t}\sum_{s} W[s,t]\,p_S[s] \notag \\
    &= \sum_{s} p_S[s] \underbrace{\sum_{t} W[s,t]}_{=1} \notag \\
    &= \sum_{s} p_S[s] = 1.
\end{align}
This ensures that P-KL produces a valid student distribution over the
\rev{teacher vocabulary when all rows sum to one. Restriction to the retained teacher support can remove probability mass, so P-KL explicitly normalizes the projected distribution on that support as in Eq.~\eqref{eq:pkl}.}

\begin{algorithm}[t]
  \caption{Rule-based projection matrix construction.}
  \label{alg:projection}
  \begin{algorithmic}[1]
    \Require Student tokenizer $\mathcal{T}_S$ with vocab $\mathcal{V}_S$;
             teacher tokenizer $\mathcal{T}_T$ with vocab $\mathcal{V}_T$;
             \rev{decay $(\beta,\gamma){=}(0.9, 0.1)$;}
             final top-$K{=}4$.
    \State \rev{Initialize empty sparse rows for the $|\mathcal V_S|\times|\mathcal V_T|$ operator $W$; do not allocate a dense matrix.}
    \Statex \textit{\# Pass 1: canonicalized exact match}
    \For{\rev{candidate string-equal pairs $(s,t)$ found by a canonical-string hash map}}
      \If{$\mathrm{canon}(\mathcal{T}_S.\texttt{decode}(s)) = \mathrm{canon}(\mathcal{T}_T.\texttt{decode}(t))$}
        \State $W[s, t] \gets 1.0$
      \EndIf
    \EndFor
    \Statex \textit{\# Pass 2: multi-token decoding rules}
    \For{each $s \in \mathcal{V}_S$ where $W[s,\cdot]$ has no exact match}
      \State $\text{text} \gets \mathcal{T}_S.\texttt{decode}(s)$
      \State $(\tau_0,\ldots,\tau_{\ell-1}) \gets \mathcal{T}_T.\texttt{encode}(\text{text})$
      \For{\rev{$i \gets 0, \ldots, \ell-1$ (full decomposition)}}
        \State $W[s, \tau_i] \gets \beta\cdot\gamma^{i}$
      \EndFor
    \EndFor
    \Statex \textit{\# Finalize: sort, truncate, row-normalize} 
    \For{each $s \in \mathcal{V}_S$}
      \State Retain only the $K$ largest entries of $W[s,\cdot]$; zero the rest.
      \State $W[s,\cdot] \gets W[s,\cdot] / \sum_j W[s,j]$
    \EndFor
    \Ensure Sparse rule-based projection matrix $W$.
  \end{algorithmic}
\end{algorithm}

\paragraph{Confidence-adaptive weight schedules.}
\label{app:confidence}
The confidence-adaptive variants in compute $\alpha_m$ from a
per-teacher confidence score derived from teacher $m$'s
predictive distribution.  For a batch with $B$ sequences of
length $N$, let
$p_{T_m}^{(b,n)}\in\mathbb{R}^{|\mathcal{V}_{T_m}|}$ denote
teacher $m$'s next-token distribution at position $n$ of batch
element $b$, and let $y^{(b,n)}\in\mathcal{V}_{T_m}$ denote the
ground-truth next token.  The three per-token confidence scores
are:
\begin{align}
  \mathrm{CE}_m^{(b,n)}
    &= -\log p_{T_m}^{(b,n)}[y^{(b,n)}],
       \label{eq:adaptive_ce}\\
  \mathrm{H}_m^{(b,n)}
    &= -\!\sum_{v\in\mathcal{V}_{T_m}}
         p_{T_m}^{(b,n)}[v]\,\log p_{T_m}^{(b,n)}[v],
       \label{eq:adaptive_entropy}\\
  \mathrm{maxp}_m^{(b,n)}
    &= \max_{v\in\mathcal{V}_{T_m}}\,p_{T_m}^{(b,n)}[v].
       \label{eq:adaptive_maxp}
\end{align}
Lower $\mathrm{CE}$ means the teacher better predicts the
ground truth; lower entropy and higher max-probability both
indicate higher teacher confidence.  We aggregate to a
per-teacher scalar by averaging over the batch and sequence
dimensions:
\begin{equation}
  \begin{split}
    \bar{w}_m &= \frac{1}{BN} \sum_{b=1}^{B} \sum_{n=1}^{N} w_m^{(b,n)}, \\
    w_m^{(b,n)} &\in \bigl\{ -\mathrm{CE}_m^{(b,n)}, -\mathrm{H}_m^{(b,n)}, \mathrm{maxp}_m^{(b,n)} \bigr\}.
  \end{split}
  \label{eq:adaptive_aggregate}
\end{equation}
where $\mathrm{CE}$ and $\mathrm{H}$ are negated so that higher
$w$ corresponds to higher teacher confidence in all three
variants.  The per-teacher mixing weights are then
\begin{equation}
  \alpha_m \;=\;
    \frac{\exp(\bar w_m)}{\sum_{m'=1}^{M}\exp(\bar w_{m'})},
  \label{eq:adaptive_softmax}
\end{equation}
producing one $(\alpha_1,\ldots,\alpha_M)$ tuple per training
step.  We also explored a per-token variant computing $\alpha_m^{(b,n)}$ per position, but observed no improvement over the per-batch formulation; the per-batch form is the default reported in our experiments.

\section{Projection Matrix Construction Details}
\label{app:projection}

\paragraph{Pseudocode for the two-pass construction.}
Algorithm~\ref{alg:projection} details the rule-based
construction of the sparse top-$4$ projection matrix $W$.
Pass~1 enumerates string-identical token pairs after
canonicalization (logically a double loop over
$\mathcal{V}_S\times\mathcal{V}_T$ as written; in practice
implemented in $O(|\mathcal{V}_S|+|\mathcal{V}_T|)$ via a
hashmap keyed on canonicalized decoded strings).  Pass~2 decodes
each student token, re-tokenizes under the teacher tokenizer,
and adds exponentially-weighted entries for each resulting
\rev{teacher sub-token. The construction in Algorithm~\ref{alg:projection} then retains the top-$K_W$ mappings and normalizes the retained row. The top-$K_W$ operation truncates a decomposition; it does not discard an entire student token whose decomposition is longer than $K_W$. P-KL additionally normalizes after projection onto the retained teacher support, while H-KL uses the top-ranked mapping. All arrays and updates use sparse storage, despite the full operator dimensions in the mathematical notation.}

\paragraph{Multi-token weight decay.}
\label{app:weight_decay}
When a student token $s$ maps to a multi-token teacher sequence
$(\tau_0, \tau_1, \ldots, \tau_{\ell-1})$ via teacher-side re-tokenization
in Pass~2, we assign weights via exponential decay and
row-normalize:
\begin{equation}
  \tilde{w}_i \;=\; \beta\cdot\gamma^{i},
  \;
  W[s, \tau_i] \;=\; \frac{\tilde{w}_i}{\sum_{j=0}^{\ell-1}\tilde{w}_j},
  \; i = 0, 1, \ldots, \ell-1,
  \label{eq:weight_decay}
\end{equation}
with \rev{$\beta = 0.9$ and $\gamma = 0.1$ by default; the additional sweep varies both.}
Explicitly, a length-2 span receives weights $(0.909, 0.091)$, a
length-3 span receives $(0.9009, 0.0901, 0.0090)$, and a
length-4 span receives $(0.9000, 0.0900, 0.0090, 0.0009)$ after
normalization.  Concentrating mass on the leading sub-token reflects the observation that it typically carries the most informative probability mass for cross-tokenizer distillation (\emph{e.g.},
``\_inter'' in [``\_inter'', ``national''] or ``\_20'' in
[``\_20'', ``24'']), while trailing sub-tokens' probability mass is less relevant
given the prefix. \rev{The additional sweep in Table~\ref{tab:init_sweep} empirically tests this initialization. Uniform weighting is worse in the aggregate, but this does not isolate numerical mappings or prove an optimal decay law.}

\paragraph{Canonicalization rules.}
The canonicalization function $\mathrm{canon}(\cdot)$ maps the decoded
string of a token to a normalized form so that functionally identical
tokens compare equal across tokenizer families.  We apply the following
rules, in order:
\begin{itemize}
  \item \textbf{Space prefix unification}: \texttt{\.{G}} (GPT-2/Llama
        BPE), \texttt{\_} (SentencePiece), and \texttt{\textvisiblespace}
        (Unicode space) all map to a single literal space character
        at the start of the token.
  \item \textbf{Newline unification}: \texttt{\.{C}}, escaped
        \texttt{\textbackslash n}, and the literal newline all map to
        \texttt{\textbackslash n}.
  \item \textbf{Byte-fallback tokens}: SentencePiece byte tokens of the
        form \texttt{<0xHH>} (for hex byte \texttt{HH}) are replaced by
        the literal character with that byte value.
  \item \textbf{Leading whitespace+punctuation pairs}: combinations
        like \texttt{\.{G},}, \texttt{\.{G}.}, \texttt{\.{G}:} are normalized to
        the punctuation alone if the combined token has an ambiguous
        whitespace interpretation.
  \item \textbf{Special tokens}: BOS, EOS, PAD, and model-specific
        chat-template tokens (\texttt{<|im\_start|>}, etc.) are
        handled separately via an explicit special-token mapping that
        pairs corresponding roles across tokenizer families (when
        unambiguous).
\end{itemize}
These rules are applied consistently at both projection-matrix
construction time and inside the DP alignment's string-equality check.
Canonicalization is idempotent and rule-based, with no learned
parameters involved.

\begin{table}[t]
  \caption{Per-category coverage check on our two cross-tokenizer
    teacher pairs: fraction of Llama tokens in each category
    surviving the 1-to-1 bijective common set $\mathcal{C}$.  The
    Qwen partition drops every multi-digit Llama numeral into
    $\mathcal{U}$; the Phi-4-mini partition keeps them all in
    $\mathcal{C}$.}
  \label{tab:audit_qwen_phi}
  \centering
  \small
  \resizebox{\linewidth}{!}{%
  \begin{tabular}{lrrr}
    \toprule
    \rev{Category} & \rev{Qwen} & \rev{Phi-4} & \rev{Size} \\
    \midrule
    1-digit numerals  & 13/13~(100\%)         & 13/13~(100\%)         & 13 \\
    2-digit numerals  & \textbf{0/100~(0\%)}  & 100/100~(100\%)       & 100 \\
    3-digit numerals  & \textbf{0/1000~(0\%)} & 1000/1000~(100\%)     & 1000 \\
    ASCII punctuation & 88/88~(100\%)         & 88/88~(100\%)         & 88 \\
    \bottomrule
  \end{tabular}
  }
\end{table}

%% file: tex/review_appendix.tex
\begin{revision}
\section{Additional Projection and Robustness Analyses}
\label{app:projection_ablations}

\paragraph{Initialization and learnability.}
Table~\ref{tab:init_sweep} evaluates initialization and sparsity. Strong decay ($\gamma=0.1$) outperforms weaker decay and uniform weighting; varying $\beta$ has little effect after normalization, and increasing support beyond four gives no aggregate gain. Learning $W$ reduces sensitivity but does not improve every configuration, supporting the rule as an empirical initialization.

\begin{table}[htbp]
\begin{revision}
\centering
\small
\setlength{\tabcolsep}{4pt}
\caption{Average accuracy (\%) for the Qwen3-4B-Base $\to$ Llama-3.2-1B projection-initialization sweep, with frozen or learnable $W$.}
\label{tab:init_sweep}
\begin{tabular}{lllrr}
\toprule
$\beta$ & $\gamma$ & $K_W$ & Frozen avg. & Learned avg. \\
\midrule
0.9 & 0.1 & 2 & 36.16 & 35.88 \\
0.9 & 0.1 & 4 & 35.91 & 35.96 \\
0.9 & 0.1 & 8 & 35.99 & 35.93 \\
0.75 & 0.1 & 4 & 35.89 & 36.02 \\
0.9 & 0.5 & 4 & 34.29 & 35.78 \\
Uniform & -- & 4 & 34.11 & 35.40 \\
0.9 & 0.9 & 4 & 34.09 & 35.40 \\
\bottomrule
\end{tabular}
\end{revision}
\end{table}

\begin{table}[htbp]
\begin{revision}
\centering
\small
\setlength{\tabcolsep}{4pt}
\caption{Qwen-to-Llama projection-matrix cosine similarity to the $(\beta,\gamma,K_W)=(0.9,0.1,4)$ reference at the reported training steps. Similarities are dimensionless; the reference is 1.000 by definition.}
\label{tab:projection_cosine}
\begin{tabular}{lllrrr}
\toprule
$\beta$ & $\gamma$ & $K_W$ & Step 0 & Step 75 & Step 100 \\
\midrule
0.9 & 0.1 & 4 & 1.000 & 1.000 & 1.000 \\
0.75 & 0.1 & 4 & 1.000 & 1.000 & 1.000 \\
0.9 & 0.1 & 8 & 0.996 & 0.996 & 0.996 \\
0.9 & 0.1 & 2 & 0.996 & 0.996 & 0.996 \\
0.9 & 0.5 & 4 & 0.928 & 0.939 & 0.939 \\
0.9 & 0.5 & 8 & 0.924 & 0.935 & 0.935 \\
0.9 & 0.9 & 4 & 0.769 & 0.795 & 0.795 \\
Uniform & -- & 4 & 0.727 & 0.753 & 0.753 \\
\bottomrule
\end{tabular}
\end{revision}
\end{table}

\paragraph{How much the projection changes.}
Table~\ref{tab:projection_cosine} shows little change when varying $\beta$ or $K_W$, with cosine similarity at least 0.996. Larger $\gamma$ and uniform weighting deviate more and recover only partially during learning, consistent with refinement of a useful initialization.

\paragraph{Long decompositions and top-$4$ truncation.}
Table~\ref{tab:decomposition_tail} reports all 443 student tokens exceeding four teacher sub-tokens. Although the $K_W=8$ sweep gives no aggregate benefit, category-specific evaluation is still needed to assess the dropped tail in non-ASCII and multilingual settings.

\begin{table}[htbp]
\begin{revision}
\centering
\small
\setlength{\tabcolsep}{4pt}
\caption{Tail of Llama-token decomposition lengths under the Qwen teacher tokenizer, over 128,256 student tokens. Exactly 443 tokens (0.35\%) have more than four teacher sub-tokens; top-$4$ retention truncates this tail.}
\label{tab:decomposition_tail}
\begin{tabular}{lr}
\toprule
Teacher sub-tokens per student token & Count \\
\midrule
5 & 286 \\
6 & 87 \\
7 & 27 \\
8 & 24 \\
$\geq9$ & 19 \\
\midrule
Total $>4$ & 443 \\
\bottomrule
\end{tabular}
\end{revision}
\end{table}

\paragraph{Seed variability.}
Across three additional matched Phi-to-Llama seeds, H-KL exceeds GOLD by 0.52 points on average, with run-to-run standard deviations of 0.12 and 0.11, respectively. The gain exceeds observed seed variation, though no formal significance test is claimed.

\end{revision}